\documentclass[10pt,letterpaper]{article}

\usepackage{cogsci}
\usepackage{pslatex}
\usepackage{amsfonts}
\usepackage{amsmath}
\usepackage{graphicx}
\usepackage{textcomp}
\usepackage{mathabx}
\usepackage{todonotes}
\usepackage{apacite}
\usepackage{bm}
\usepackage{enumitem}
\usepackage{mathtools}

\newcommand\states{\mathcal{S}}
\newcommand\actions{\mathcal{A}}
\newcommand\obs{\mathcal{Z}}
\newcommand\tfunc{\mathcal{T}}

\newcommand\agents{\mathcal{I}}
\newcommand\expectation{\mathbb{E}}
\newcommand\vecsup[2]{\vec{#1}^{\,#2}}

\newenvironment{myitemize}
{ \begin{itemize}
    \setlength{\itemsep}{0pt}
    \setlength{\parskip}{0pt}
    \setlength{\parsep}{0pt}     }
{ \end{itemize}                  } 

\title{The Computational Structure of Unintentional Meaning}

\author{
  {\large \bf Mark K. Ho (mho@princeton.edu)} \\
  Department of Psychology, Princeton University \\
  Princeton, NJ 08540 
    \vspace{-2mm}
    \\
  \AND
  {\large \bf Joanna Korman\footnotemark (jkorman@mitre.org)} \\
  The MITRE Corporation \\
  Bedford, MA 01730 
      \vspace{-2mm}
  \AND
  {\large \bf Thomas L. Griffiths (tomg@princeton.edu)}\\
  Department of Psychology, Princeton University \\
  Princeton, NJ 08540
  }

\begin{document}
\maketitle
  
\begin{abstract}
Speech-acts can have literal meaning as well as pragmatic meaning, but these both involve consequences typically \textit{intended} by a speaker. Speech-acts can also have \textit{unintentional meaning}, in which what is conveyed goes above and beyond what was intended. Here, we present a Bayesian analysis of how, to a listener, the meaning of an utterance can significantly differ from a speaker's intended meaning. Our model emphasizes how comprehending the intentional \textit{and} unintentional meaning of speech-acts requires listeners to engage in sophisticated model-based perspective-taking and reasoning about the \textit{history} of the state of the world, each other's actions, and each other's observations. To test our model, we have human participants make judgments about vignettes where speakers make utterances that could be interpreted as intentional insults or unintentional \textit{faux pas}. In elucidating the mechanics of speech-acts with unintentional meanings, our account provides insight into how communication both functions and malfunctions.

\end{abstract}

\section{Introduction}
\footnotetext{\scriptsize The author's affiliation with The MITRE Corporation is provided for identification purposes only, and is not intended to convey or imply MITRE's concurrence with, or support for, the positions, opinions, or viewpoints expressed by the author.}
People sometimes communicate things that they did not intend or expect. Consider the following vignette, adapted from~\shortciteA{baroncohen1999recognition}:
\begin{quote}
\textbf{Curtains} Paul had just moved into a new apartment. Paul went shopping and bought some new curtains for his bedroom. After he returned from shopping and had put up the new curtains in the bedroom, his best friend, Lisa, came over. Paul gave her a tour of the apartment and asked, ``How do you like my bedroom?'' 

``Those curtains are horrible,'' Lisa said. ``I hope you're going to get some new ones!''
\end{quote}

Clearly, Lisa committed a social blunder or \textit{faux pas} with her remark. What happened here? When Lisa says, ``Those curtains look horrible,'' she is merely stating her private aesthetic experience of the curtains. The \textit{literal} meaning is straightforward: The curtains look bad. And the \textit{intended} or \textit{expected} meaning of her utterance is largely captured by this literal meaning. However, to Paul, the utterance means more. Specifically, what Lisa is \textit{really} saying is that \textit{he} chose horrible curtains. Of course, Lisa did not ``really'' say that Paul's choice in curtains was horrible---she had no intention of conveying such an idea. Paul might even realize this. Nonetheless, the remark stings. Why? Lisa and Paul each possess a piece of a puzzle, and when put together, they entail that Paul has awful taste in curtains. At the outset, neither one knew that they each had a piece of a puzzle. But once Lisa makes her remark, she inadvertently completes the puzzle, at least from Paul's perspective.

Standard models of communication~\cite{grice1957meaning, sperber1986} tend to focus on how people use language successfully. For example, people can imply more than they literally mean~\cite{carston2002thoughts}, convey subtle distinctions via metaphor~\cite{tendahl2008complementary}, and manage their own and others' public face using politeness~\cite{levinson1987politeness, yoon_frank_tessler_goodman_2018}. But things do not always go smoothly, as Paul and Lisa's situation indicates. Sometimes people find themselves having inadvertently stepped on conversational landmines, meaning things that they never anticipated meaning. Notably, because such situations present complex dilemmas of mutual perspective-taking against a backdrop of divergent knowledge, they can serve as advanced tests of theory of mind~\cite{baroncohen1999recognition, Zalla2009, korman2017action}. But how do people reason about such dilemmas? And how can this be understood computationally? Disentangling unintentional meaning can shed light on how communication works in a broader social context as well as inform the design of artificial intelligences that interact with people.

Here, we develop a rational, cognitive account of interpreting unintentional speech-acts that builds on existing Bayesian models of language (e.g., Rational Speech Act [RSA] models [\citeNP{goodman2016pragmatic}]). To do this, we analyze the general epistemic structure of social interactions such as the one described above and model listeners engaging in \textit{model-based perspective-taking}. In particular, our model explains how the same utterance could be interpreted as either an (unintentional) \textit{faux pas} or an \textit{intentional insult} depending on the context of a listener and speaker's interaction. We then test several model predictions in an experiment with human participants. In the following sections, we outline our computational model, experimental results, and their implications.

\section{A Bayesian Account of Unintentional Meaning}

\begin{figure*}[ht]
\begin{center}
\includegraphics[width=.92\textwidth,height=\textheight,keepaspectratio]{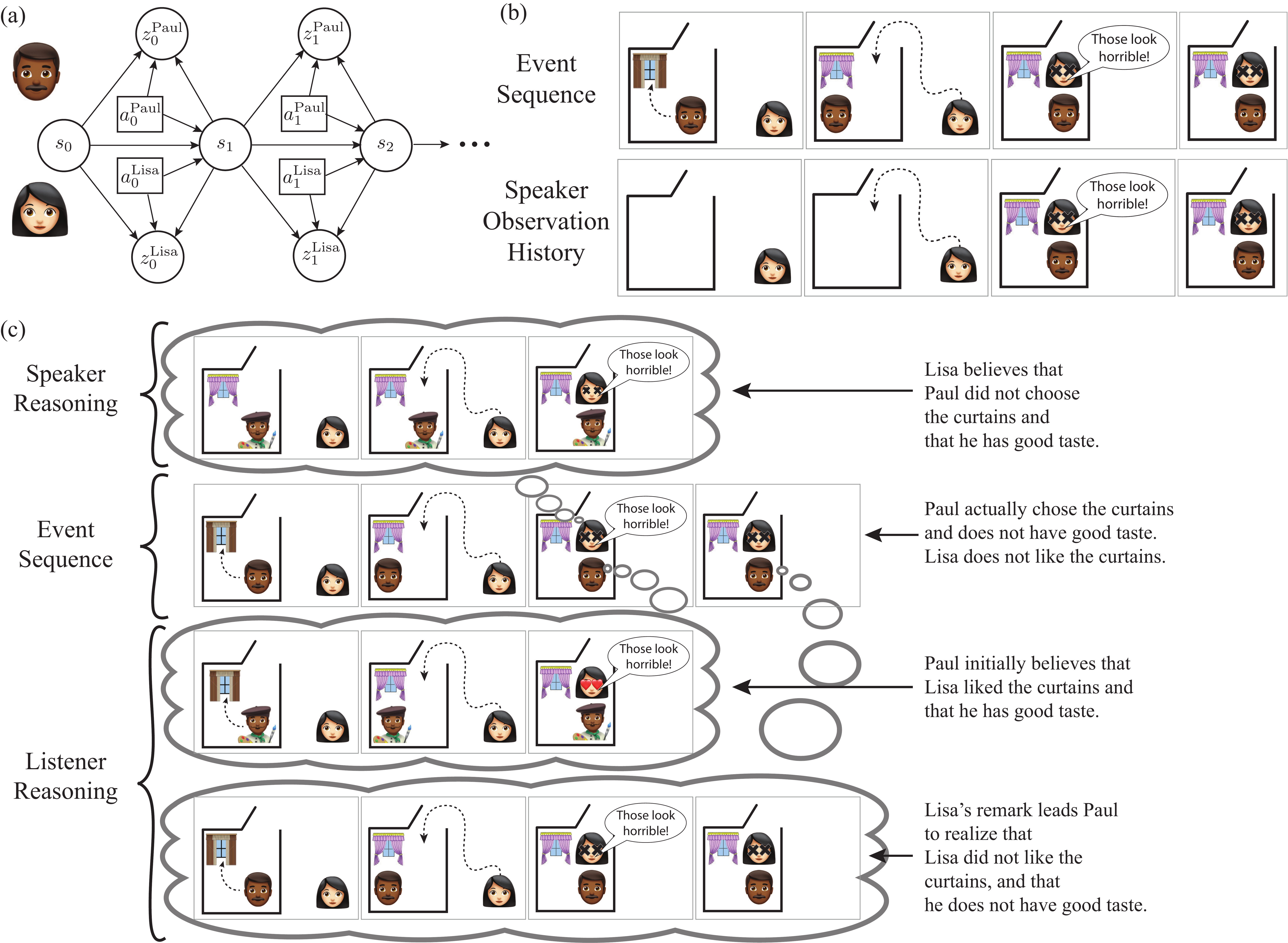}
\end{center}
\vspace{-4mm}
\caption{Model and example of unintentional meaning. (a) Influence diagram with state, action, and observation dependencies. Circles correspond to world state (e.g., $s_t$) and observation (e.g., $z_t^{i}$) variables; squares correspond to agent action variables (including utterances) (e.g., $a_t^i$). (b) Event sequence in \textbf{Curtains} (top) and speaker observation history (bottom). Lisa does not observe Paul choose the curtains. Only Lisa experiences whether the curtains look good or bad and comments on this experience. (c) Diagram of interactive belief state over time in \textbf{Curtains}. 
}
\label{fig:model}
\vspace{-2mm}
\end{figure*}

During social interactions, people reason about the world as well as each other's perspective on the world~\cite{brown2018perspective}. Thus, our account has two components, which we formulate as probabilistic models. First, we specify a \textit{world model} that captures common-sense relationships between world states, actions, and events. Second, we define \textit{agent models} of a speaker and listener reasoning about the world and one another.

\subsection{World Model}
We model the interaction as a partially observable stochastic game (POSG), a generalization of Markov Decision Processes (MDPs) with multiple agents with private observations~\cite{kuhn1953extensive}. Formally, a world model $\mathcal{W} = \langle \agents, \states, \actions, \obs, \tfunc \rangle$ where:
\vspace{-2mm}
\begin{myitemize}
\item $\agents$ is a set of $n$ agents indexed $1, ..., n$; 
\item $\states$ is a set of possible states of the world, where each state $s \in \states$ is an assignment to $k$ variables, $s = (x_0, x_1, ..., x_k)$; 
\item $\actions = \bigtimes_{i \in \agents} \actions^i$ is the set of joint actions, i.e., every combination of each agent $i$'s actions, $\actions^i$ (including utterances);
\item $\obs = \bigtimes_{i \in \agents} \obs^i$ is the set of joint private observations, which is every possible combination of each individual agent $i$'s private observation set, $\obs^i$; and
\item $\tfunc = P(z, s' \mid s, a)$ is a transition function representing the probability of a joint observation $z$ and next state $s'$ given a previous state $s \in \states$ and joint action $a \in \actions$ was taken.
\end{myitemize}
\vspace{-1mm}

In \textbf{Curtains}, the initial state, $s_0$, includes Paul with the old curtains in the apartment and Lisa elsewhere. There is also a latent state feature of interest: whether Paul has good or bad taste. At $t=0$, Paul's action, $a^{\text{Paul}}_0$, is choosing new curtains, while Lisa's action, $a^{\text{Lisa}}_0$, is going to the apartment. The joint action, $a_0 = (a^{\text{Paul}}_0, a^{\text{Lisa}}_0)$, results in a new state, $s_1$, with them both in the apartment, the curtains either good or bad, and Paul's taste. Paul's observation, $z^{\text{Paul}}_0$, but not Lisa's, $z^{\text{Lisa}}_0$, includes Paul having put up the curtains. These relationships between world states (e.g. Paul and Lisa's locations), actions (e.g. Lisa walking to Paul's apartment), and observations (e.g. Paul observing himself put up the curtains) are formally encoded in the transition function $\tfunc$. The sequence of states, joint actions and observations resulting from such interactions constitute the \textit{history} up to a point $t$, $\vec{h_{t}} = (s_0, a_0, z_0, ..., s_{t-1}, a_{t-1}, z_{t-1}, s_t)$.

\subsection{Agent Models}
Agents are modeled as Bayesian decision-makers~\cite{bernardo1994bayesian} who can reason about the world and other agents as well as take actions---including making utterances. 

\subsubsection{Interactive Belief State}
Agents' beliefs are probability distributions over variables that represent aspects of the current state, previous states, or each other's beliefs. The configuration of these first- and higher-order, recursive beliefs constitute their \textit{interactive belief state}~\cite{gmytrasiewicz2005framework}. We refer to an agent $i$'s beliefs as $b^i$. For example, if we denote Paul's taste as the variable $T^{\text{Paul}}$, then Paul's belief that his taste is good is $b^{\text{Paul}}(T^{\text{Paul}} = \texttt{Good})$. Higher-order beliefs can also be represented. For instance, we can calculate Paul's \textit{expectation} of Lisa's belief in his taste as $\expectation_{b^{\text{Paul}}} [b^{\text{Lisa}}](T^{\text{Paul}})  = \sum_{b^{\text{Lisa}}}b^{\text{Paul}}(b^{\text{Lisa}}(T^{\text{Paul}}))$. 

An agent $i$'s beliefs are a function of their prior, model of the world, model of other agents, and \textit{observation history} up to time $t$, $\vecsup{z}{i}_{t}$. Note that $\vecsup{z}{i}_{t}$ can include observations that are completely private to $i$ (e.g., Lisa's personal aesthetic experience) as well as public actions and utterances (e.g., Lisa's remark to Paul). Thus, we denote Paul's belief about his taste at a time $t$ as $b^{\text{Paul}}_t(T^{\text{Paul}}) = b^{\text{Paul}}(T^{\text{Paul}} \mid \vecsup{z}{\text{Paul}}_{t})$. Given a sequence of observations, $\vecsup{z}{i}_{t}$, posterior beliefs about a variable $X$ are updated via Bayes' rule:
\begin{align}
b(X \!\mid \! \vecsup{z}{i}_{t})& \ \propto \  b(\vecsup{z}{i}_{t}\!\mid\!X) b(X) \\
&= \sum_{\vec{h}_t} b(\vecsup{z}{i}_{t}\!\mid\!\vec{h}_t)b(\vec{h}_t, X)
\label{eq:history_reasoning}
\end{align}

The capacity to reason about higher-order beliefs (e.g., Paul's beliefs about Lisa's belief in his taste), along with Equation~\ref{eq:history_reasoning} express agents' joint inferences about events and model-based perspective-taking.

\subsubsection{Speaker Model}
Speakers have beliefs and goals. When choosing what to say, they may have beliefs and goals with respect to the listener's beliefs and goals. In our example, Lisa may care about being informative about how she sees the curtains, but may also think Paul cares about having good taste in curtains and care whether she hurts his feelings. Following previous work (e.g., \citeNP{franke2009signal}), we model speakers as reasoning about changes in \textit{belief states}. Here, we are interested in how a speaker can intend to mean one thing but inadvertently mean another. Thus, we distinguish between state variables that the speaker wants to be \textit{informative} about, $X^{\text{Info}}$ (e.g., how Lisa sees the curtains), and \textit{evaluative} variables, $X^{\text{Eval}}$, that the listener wants to take on a specific value $x^{\text{Eval}*}$ (e.g., Paul's taste being good). The speaker then cares about the changes in those quantities. Formally:
\begin{equation}
\label{eq:delta_info}
\Delta^{\text{L-Info}}_t = b^L_{t+1}(X^{\text{Info}} = x^{\text{Info}}) - b^L_t(X^{\text{Info}} = x^{\text{Info}}),
\end{equation}
where $x^{\text{Info}}$ is given by $\vec{h}_{t}$; and,
\begin{equation}
\label{eq:delta_eval}
\Delta^{\text{L-Eval}}_t = b^L_{t+1}(X^{\text{Eval}} = x^{\text{Eval}*}) - b^L_t(X^{\text{Eval}} = x^{\text{Eval}*}).
\end{equation}

A speaker who is interested in what the listener thinks about $X^{\text{Info}}$ and $X^{\text{Eval}}$ will, at a minimum, anticipate how their utterances will influence $\Delta^{\text{L-Info}}_t$ and $\Delta^{\text{L-Eval}}_t$. A speaker would then have a reward function defined as:
\begin{equation}
\label{eq:speaker_reward}
R^{\text{S}}(a_t^S, \vecsup{z}{L}_{t+1}) = \theta^{\text{L-Info}}\Delta^{\text{L-Info}}_t + \theta^{\text{L-Eval}}\Delta^{\text{L-Eval}}_t
\end{equation}
where the $\theta$ terms correspond to how the speaker values certain outcomes in the listener's mental state. For instance, if $\theta^{\text{L-Eval}} < 0$, the speaker \textit{wants} to insult the speaker. 

Given Equation~\ref{eq:speaker_reward}, a speaker can take utterances based on expected future utility/rewards (or \textit{value} [\citeNP{sutton_reinforcement_1998}]), where the expectation is taken with respect to the speaker's beliefs, $b^S_t$. That is, given observations $\vecsup{z}{S}_{t}$, the value of $a^S_t$ is
$
V^S(a^S_t; \vecsup{z}{S}_{t}) = 
\mathbb{E}_{b^S_t}
\big[
R^{\text{S}}(a_t^S, \vecsup{z}{L}_{t+1})
\big]
$
,
and an action is chosen using a Luce choice rule~\cite{luce1959possible}.

\subsubsection{Listener Inference}
Our goal is to characterize how a listener's interpretation of an utterance can differ from a speaker's intended meaning, which requires specifying listener inferences. We start with a simple listener that understands the literal meanings of words when spoken. Following previous models \cite{franke2009signal, goodman2016pragmatic}, the literal meaning of an utterance $a^S$ is determined by its truth-functional denotation, which maps histories to Boolean truth values, $[\![a^S]\!]: \vec{h}_t \mapsto y$, $y \in \{\mathtt{True}, \mathtt{False}\}$. A literal listener's model of speaker utterances is:
\begin{equation*}
b(a^S \mid \vec{h}_t) \ \propto \ 
\begin{cases}
      1 - \varepsilon & \text{if }[\![a^S]\!](\vec{h}_t) \\
      \varepsilon & \text{if } \lnot[\![a^S]\!](\vec{h}_t)\\
    \end{cases} 
\end{equation*}
where $\varepsilon$ is a small probability of $a^S$ being said even if it happens to be false.

We can also posit a more sophisticated listener who, rather than assuming utterances literally reflect reality, reason about how a speaker's beliefs and goals mediate their use of language. This type of listener draws inferences based on an \textit{intentional} model of a speaker that track the quantities in Equations~\ref{eq:delta_info} and \ref{eq:delta_eval} as well as maximize the expected rewards. These inferences, however, occur while the listener is also reasoning about the actual sequence of events $\vec{h}_t$, making it possible to draw inferences based on utterances that the speaker did not anticipate.

\section{Model Simulations}
\begin{figure*}[!ht]
\begin{center}
\includegraphics[width=\textwidth,height=\textheight,keepaspectratio]{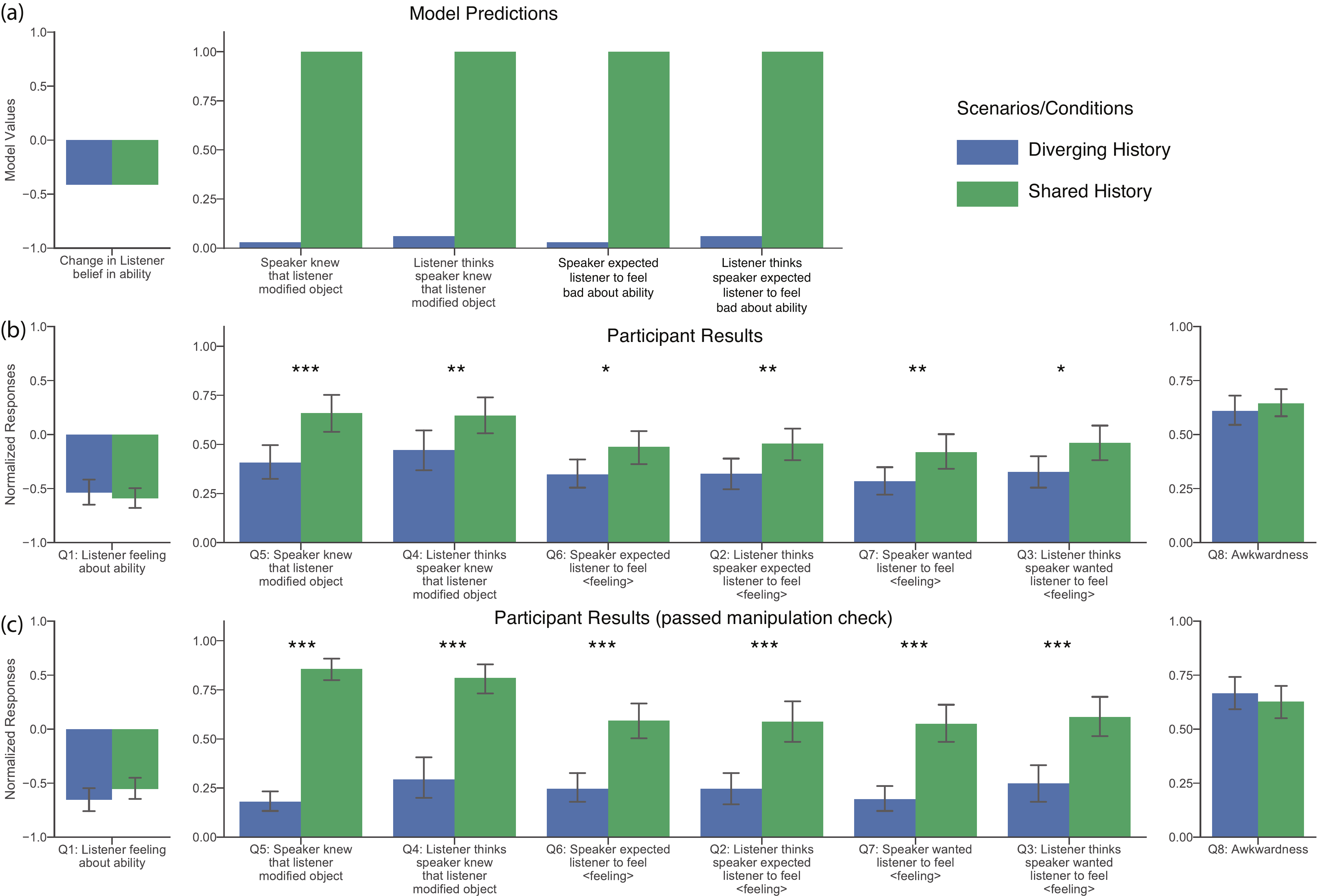}
\end{center}
\vspace{-4mm}
\caption{(a) Model predictions. The model predicts that the listener's change in belief in the evaluative variable ($\Delta^{\text{L-Eval}}_t$) is equally negative in the diverging and shared history scenarios. However, whether the speaker anticipated the offensiveness of their comment differs between the two scenarios, as do the listener's beliefs about the speaker's anticipation. (b) Judgments from all participants by question. Responses were normalized depending on whether response scales were valanced (Q1), likelihood (Q2-Q7), or qualitative (Q8). (c) Judgments from participants who correctly identified whether the speaker knew the listener modified the object. ${}^{*}: p < .05$, ${}^{**}: p < .01$, ${}^{***}: p < .001$.}
\label{fig:simexp}
\end{figure*}

In the original \textbf{Curtains} scenario, Lisa was not present when Paul put up the curtains. As a result, Lisa's comment (``Those curtains are horrible'') is interpreted in a \textit{diverging} observation history context. But what if Lisa had been present when Paul put up the curtains and made the same utterance? Given a \textit{shared} observation history, Lisa's utterance is still offensive, but now Lisa has all the information needed to realize it would be offensive. Put simply, in the \textit{diverging history} context, the utterance is a faux pas, whereas in the \textit{shared history} context, it is an intentional insult.

In this section, we discuss how our model can be used to make these intuitive predictions precise and explain how they arise from agents' interactions and model-based perspective-taking within a shared environment. We implemented our model in WebPPL~\cite{goodman_dippl}, a programming language that can express stochastic processes like POSGs as well as Bayesian inference. 

\subsection{Generative Model}

To model a scenario like \textbf{Curtains}, we define agents, objects, and features assigned to them. These are the curtains, which have a location (inside Paul's apartment); the speaker (Lisa), who has a location (inside or outside Paul's apartment) and a perception of the curtains (good or bad); and the listener (Paul), who has a location (inside or outside) and ability to choose curtains (high or low). Additionally, the listener can either act on the curtains or not, while the speaker can enter the apartment and make an utterance about the curtains (``the curtains look good'', ``the curtains look bad'', or \texttt{<nothing>}). The truth-conditional semantics of the utterances map onto world features in a standard manner, and we set $\varepsilon = .05$.

Observations depend on whether agents and objects are co-located and are defined as subsets of state and action variables. For instance, if Paul and Lisa are both inside the house and Paul modifies the curtains, they both observe that Paul acted on the curtains, but only Lisa directly knows whether they look good to her. Finally, we define a state and action prior for both agents such that the listener's ability is initially high ($p = 0.90$), the speaker's perception of the object is initially random ($p = 0.50$), and the listener has a low probability of modifying the object ($p = 0.05$). 

\subsection{Model Predictions}

Given the generative model, we can provide scenarios and calculate aspects of the resulting interactive belief state (the listener and speaker's beliefs about the world and each other's beliefs). In particular, we compare the results of a \textit{shared history} with those of a \textit{diverging history}. In the shared history, the speaker and listener are both present when the listener modifies the object, whereas in the diverging history, the speaker is not present when the listener acts on the object. Otherwise, the two scenarios are the same and the speaker comments on the curtains being bad. Figure~\ref{fig:simexp}a displays the results of the simulation when given each of the two histories. In both histories, the listener learns that their ability when modifying the object, $X^{\text{Eval}}$, is low (i.e., $\Delta^{\text{L-Eval}}_t < 0$). They also learn about the informative variable (i.e., $\Delta^{\text{L-Info}}_t > 0$).

However, the resulting interactive belief states differ in important ways. For example, in the diverging history, although the listener concludes that the evaluative variable is \textit{low}, the speaker thinks the evaluative variable is \textit{high}. Relatedly, the speaker thinks the utterance was informative ($\expectation_{b^S}[\Delta^{\text{L-Info}}] > 0$) but not offensive ($\expectation_{b^S}[\Delta^{\text{L-Eval}}] = 0$). Moreover, the listener knows the speaker believes that their comment was expected to be informative and not offensive. In the shared history, this is not the case: The listener and speaker both believe the evaluative variable is low, and they both know the resulting informational and evaluative effects. Because they were both present when the listener modified the object, they share expectations about the utterance's meaning. 

Put intuitively, whereas the shared history leads to an \textit{expected insult}, the diverging history leads to a \textit{faux pas}. Our model explains this difference in terms of differential transformations of the listener and speaker's interactive belief state.

\section{Experiment}
Our model explains how different observation histories result in interactive belief states, which can produce unintentional meaning. To test whether this accurately describes people's capacity to reason about unintentional meaning, we had people read vignettes that described scenarios involving shared or diverging observation histories. The underlying logical structure of all the vignettes mirrored that of \textbf{Curtains}, and so the model predictions described in the previous section apply to all of them. Participants then provided judgments corresponding to predicted differences in listener/speaker beliefs. The study's main hypotheses were preregistered on the Open Science Framework platform (\texttt{https://osf.io/84wqn}). Overall, we find that our model captures key qualitative features of people's inferences.

\subsection{Materials}
We developed a set of vignettes that included interactions in different contexts as well as different histories of interaction. Each vignette involved a listener (e.g., Paul) who could potentially interact with an object (e.g., curtains) as well as a speaker (e.g., Lisa) who makes an utterance about their negative aesthetic experience of the object (e.g., ``The curtains look horrible''). In the shared history versions of the vignettes, the two agents were described as being both present when the listener acted on an object. In the diverging history versions of the vignettes, the speaker was not present when the listener interacted with the object. Each vignette involved one of five contexts: \textit{Curtain}, \textit{Story-Prize}, \textit{Wine-bottle}, \textit{Cupcakes}, and \textit{Parking}. Thus there were a total of ten items (Diverging/Shared history $\times$ 5 contexts). All items used in the experiment are available on the primary author's website.

\subsection{Procedure}
One-hundred participants were recruited via MTurk to participate in our experiment using PsiTurk~\cite{gureckispsiturk2016}. Each participant read one of the ten context-history items, and then answered the following questions in order:
{
\newcommand\Lis{\texttt{<listener>}}
\newcommand\Sp{\texttt{<speaker>}}
\newcommand\Act{\texttt{<action>}}
\newcommand\Qresp{\texttt{<Q1\_response>}}
\newcommand\rscale{[Definitely Not - Definitely]}
\begin{itemize}[noitemsep,nolistsep]
\item{Q1: At this point, how does \Lis{} feel about their ability to \Act? [6 point scale ranging ``Very Bad'' to ``Very Good'' with no neutral option]}
\item{Q2: \Lis{} thinks that \Sp{} expected that their remark would make them feel \Qresp.}
\item{Q3: \Lis{} thinks that in making the remark, \Sp{} wanted to make them feel \Qresp.}
\item{Q4: \Lis{} thinks that \Sp{} thinks that \Lis{} \Act{}.}
\item{Q5: \Sp{} knew that \Lis{} \Act{}.}
\item{Q6: In making the remark, \Sp{} expected \Lis{} to feel \Qresp.}
\item{Q7: In making the remark, \Sp{} wanted \Lis{} to feel \Qresp.}
\item{Q8: How awkward is this situation? [5 point scale ranging ``Not at all'' to ``Extremely''}
\end{itemize}
The values for \Lis, \Sp, and \Act were specified parametrically based on the context, while the value for \Qresp{} was filled in based on the answer to the first question. The response scale for questions 2-7 was a six-point scale ranging from ``Definitely Not'' to ``Definitely'', with no neutral point. We included question 8 because previous work studying faux pas have focused on this question~\cite{Zalla2009}. Participants were also given free response boxes to elaborate on their interpretation of the situation and answered demographic questions.
}

\begin{table}[!hb]
\begin{center}
\begin{tabular}{cccccl}
\hline
Question & $\beta$ &  S.E. &     $d\!\!f$ & $t$ & $p$ \\
\hline
Q1 & -0.06 &  0.07 &  94.0 & -0.77 &       \\
Q2 &  0.15 &  0.06 &  94.0 &  2.65 &    ** \\
Q3 &  0.15 &  0.06 &  94.0 &  2.50 &     * \\
Q4 &  0.18 &  0.06 &  94.0 &  2.78 &    ** \\
Q5 &  0.25 &  0.06 &  94.0 &  4.34 &   *** \\
Q6 &  0.14 &  0.06 &  94.0 &  2.53 &     * \\
Q7 &  0.15 &  0.06 &  94.0 &  2.64 &    ** \\
Q8 &  0.04 &  0.05 &  94.0 &  0.78 &       \\
\hline
\label{table:linear_res}
\end{tabular}
\end{center}
\vspace{-7mm}
\caption{Tests for Diverging/Shared history factor.}
\vspace{-2mm}
\end{table}

\subsection{Experimental Results}
\newcommand\lme[5]{$\beta = #1$, SE$= #2$, $t(#3) = #4$, $p #5$}
\subsubsection{Manipulation check} To assess whether the Diverging/Shared history manipulation worked, we examined responses to Q5 (whether the speaker knew the listener acted on the object). A comparison in which the responses were coded as Yes or No (i.e., above or below the middle of the response scale) showed that it was effective ($\chi^2(1) = 7.92, p < .01$). However, a number of participants ($15$ of $50$ in Shared; $20$ of $50$ in Diverging) did not pass this manipulation check and gave opposite answers than implied by the stories. Whether their responses are included does not affect our qualitative results, and in our analyses we use the full data set. Figure~\ref{fig:simexp}c plots the results for those who passed this check.

\begin{figure}[t]
\begin{center}
\includegraphics[width=.45\textwidth,keepaspectratio]{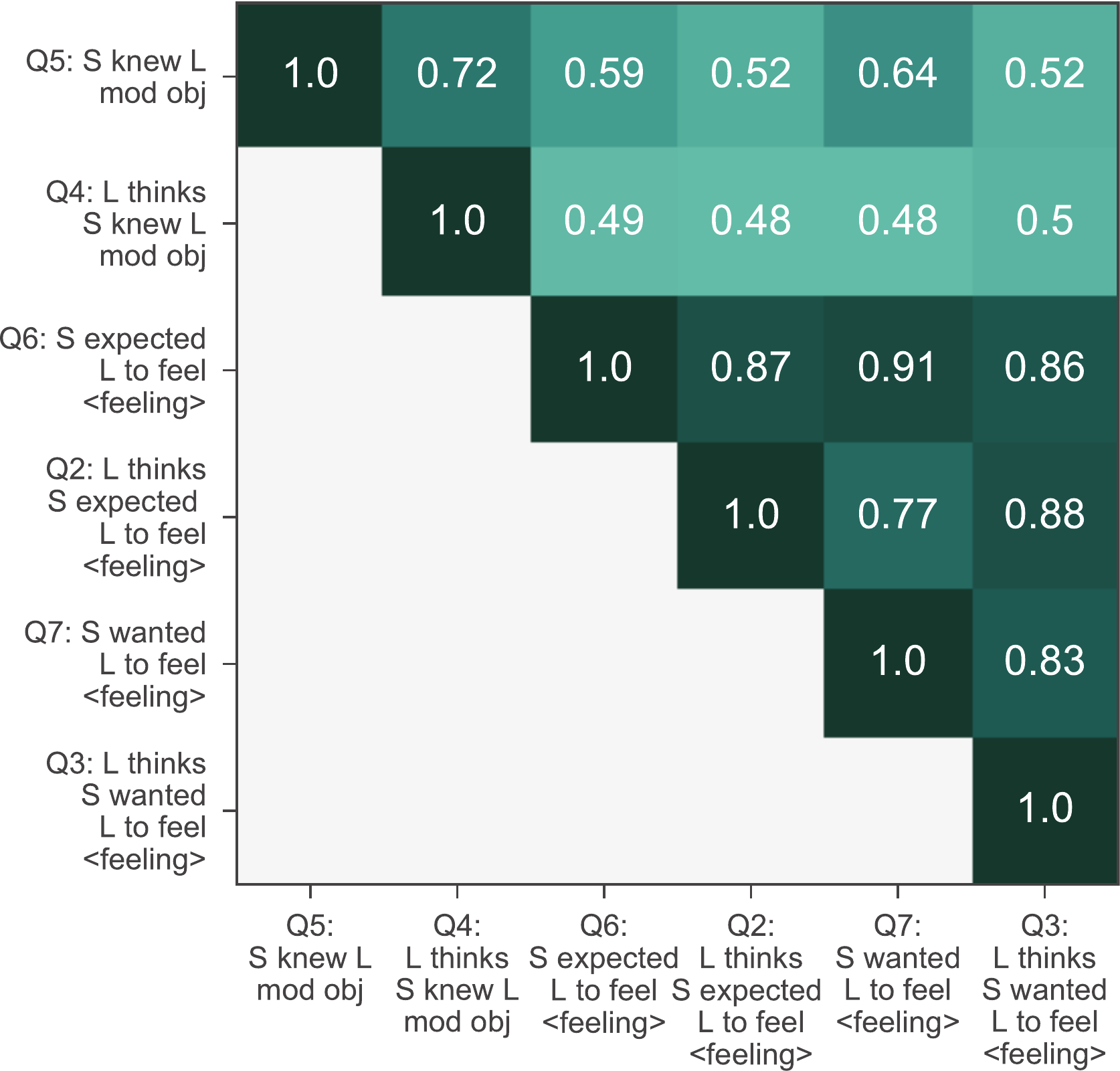}
\end{center}
\vspace{-5mm}
\caption{Judgment correlations (Pearson's $r$).}
\label{fig:judgment_corrs}
\vspace{-4mm}
\end{figure}

\subsubsection{Judgment differences} Responses paralleled the model predictions for the Shared versus Diverging history versions of the vignettes (Figure~\ref{fig:simexp}b). For each judgment, we fit mixed-effects linear models with context intercepts as a random effect and history as a fixed effect. Table~\ref{table:linear_res} shows tests of significance on the Diverging/Shared history parameters. Judgments about the listener's feelings (Q1) were negative and not significantly different, indicating that people perceived the psychological impact (at least with respect to ability) of the utterance as roughly equivalent. In contrast, questions about the interactive belief state---the listener and speaker's beliefs about the world and each other's beliefs (Q2-Q7)---differed as predicted by the model. In particular, participants thought that the speaker neither expected that their utterance would hurt the listener's feelings, nor that they wanted to do so. Participants judged that the listener recognized this as well.

\subsubsection{Judgment correlations} Judgments among questions about higher order mental states were strongly correlated, while those between the higher order mental states and the listener's action were weaker (Figure~\ref{fig:judgment_corrs}). Specifically, those about speaker mental states (Q6, Q7) and listener beliefs about speaker mental states (Q2, Q3) were all highly correlated (all $r \in [0.77, 0.91], p < .001$). In contrast, questions about knowledge of the object being modified (Q4, Q5) were only moderately correlated with those about anticipated effects (Q2, Q3, Q6, Q7) (all $r \in [0.48, 0.64], p < .001$).

\section{Discussion}
People's actions can have unexpected consequences, and speech-acts are no different. To understand unintentional meaning though, we need to characterize how a communicative act can lead to unanticipated epistemic consequences. Sometimes, a listener can learn something from an utterance that a speaker did not intend to convey or may not even believe (e.g., as in \textbf{Curtains}). Here, we have presented a Bayesian model and experiments testing how people reason about scenarios involving unintentional speech acts. Specifically, our account treats speech-acts as actions taken by a speaker that influence a shared interactive belief state---the beliefs each agent has about the world and each other's beliefs. In doing so, we can capture the inferences that underlie unintentional meaning.

The current work raises important empirical and theoretical questions about how people reason about interactive beliefs and unintentional meaning. For instance, our experiments focus on third-party judgments about how a listener interprets the unintended meanings of utterances, but further work would be needed to assess how listeners do this (e.g., when the victim of an offhand comment) or even how speakers can recognize this (e.g., realizing one has put their foot in their mouth). Additionally, we have presented a Bayesian account of unintentional meaning in which agents reason about a large but finite set of possible histories of interaction. In everyday conversation, the space of possible histories can be much larger or even infinite. It is thus an open question how people can approximate the recursive inferences needed to make sense of unintentional meaning.

A rigorous characterization of unintentional meaning can deepen our understanding of how communication works in a broader social context. For example, attempts to build common ground through shared experience~\cite{clark1981definite, mckinley2017memory} or manage face with polite speech~\cite{levinson1987politeness, yoon_frank_tessler_goodman_2018} could be understood, in part, as strategies for forestalling unintentional meaning. And given that intentionality plays a key role in judgments of blame~\cite{baird2004}, phenomena like plausible deniability could be understood as people leveraging the possibility of unintentional meaning to covertly accomplish communicative goals~\cite{pinker2008logic}. Although further investigation is needed to test the extent to which people can track and influence interactive belief states (as well as how artificial agents can do so), this work provides a point of departure for computationally investigating these social and cognitive aspects of communication.

\section{Acknowledgments}
This material is based upon work supported by the NSF under Grant No. 1544924.
\newpage

\setlength{\bibleftmargin}{.125in}
\setlength{\bibindent}{-\bibleftmargin}
{
\renewcommand{\bibliographytypesize}{\footnotesize} 

}

\end{document}